%% file: main.tex
\documentclass[journal]{IEEEtran}

\ifCLASSINFOpdf
\else
\fi
\usepackage{booktabs}
\usepackage{cite}

\usepackage{overpic}
\usepackage{bbding}
\usepackage{rotating}

\usepackage{booktabs}
\usepackage{amsmath}
\usepackage{multirow}
\usepackage{multicol}
\usepackage{colortbl}
\usepackage{amssymb}

\usepackage[table]{xcolor}
\usepackage{subfigure}
\usepackage{overpic}

\usepackage{xcolor}
\usepackage{pgfplots}

\usepackage{pgfplots}
\usepgfplotslibrary{groupplots}
\usetikzlibrary{patterns}
\usepackage{tikz}
\definecolor{myblue}{HTML}{92dcec}

\definecolor{hous}{HTML}{b88b4d}
\definecolor{green}{HTML}{79c561}
\definecolor{farming}{HTML}{ded94c}
\definecolor{trans}{HTML}{b4b4a9}
\definecolor{services}{HTML}{ff362e}
\definecolor{other}{HTML}{dbd4d3}
\definecolor{industry}{HTML}{db79c0}
\definecolor{water}{HTML}{7982db}
\definecolor{techinfra}{HTML}{303355}
\definecolor{aliceblue}{rgb}{0.94, 0.97, 1.0}

\usepackage[pagebackref,breaklinks,colorlinks]{hyperref}
\usepackage[capitalize]{cleveref}

\crefname{section}{Sec.}{Secs.}
\Crefname{section}{Section}{Sections}
\crefname{table}{TABLE}{TABLES}

\usepackage{nicematrix}
\usepackage{tikz}
\usetikzlibrary{fit}
\graphicspath{{figures/}}
\usepackage{booktabs} 
\usepackage{multirow} 
\usepackage{xcolor} 

\begin{document}
%
\title{\textbf{IMPACT}: \textbf{I}mage-Conditioned \textbf{M}otion with \textbf{P}hysics-\textbf{A}ware Collision \\and \textbf{T}rajectory Guidance for Video Generation}


%
%


\author{Shiwei Yang, Yunlong~Yu$^\dagger$, Zhao~Wang, and Jungong Han
\thanks{$^\dagger$ is the corresponding author.}
\thanks{Shiwei Yang and Yunlong Yu are with the College of Information Science and Electronic Engineering, Zhejiang University, Hangzhou, 310027 China (e-mail: \{shiweiyang, yuyunlong\}@zju.edu.cn).}
\thanks{Zhao Wang is with Ningbo Innovation Center, Zhejiang University, Ningbo, China (e-mail: zhao\_wang@zju.edu.cn). }
\thanks{Jungong Han is with the Department of Automation, Tsinghua University, Beijing 100084 (e-mail: jungonghan77@gmail.com).}
}

%



\maketitle

\input{Sections/0-abstract}

\begin{IEEEkeywords}
Physics-consistent video generation,
image-to-video generation,
physics-guided motion,
trajectory guidance.
\end{IEEEkeywords}

\IEEEpeerreviewmaketitle

\input{Sections/1-introduction}

\input{Sections/2-relatedwork}

\input{Sections/3-method}
\input{Sections/4-experiments}
\input{Sections/5-conclusion}

\section*{Acknowledgment}
This work was supported in part by the National Natural Science Foundation of China under Grant 62576313, the Key Research and Development Program of Zhejiang Province under Grant 2026C01016, and the Key Research and Development Program of Ningbo under Grant 2025Z128.

\ifCLASSOPTIONcaptionsoff
  \newpage
\fi

\bibliographystyle{IEEEtran}
\bibliography{references}

\end{document}

%% file: Sections/0-abstract.tex
\begin{abstract}
Visually realistic video generation does not guarantee physically consistent motion. Generated objects may exhibit implausible collision responses or rebounds without energy dissipation, while even physically valid motion predictions may not be faithfully realized during synthesis. We propose \textbf{IMPACT}, a physics-state-driven image-to-video framework that addresses these challenges by coupling explicit dynamics prediction with complementary sparse--dense physical guidance. A learnably calibrated Physics State Generator unifies continuous free-flight dynamics with discrete collision responses, linking impact timing, post-impact velocity, rebound attenuation, and eventual settling within a coherent physical-state sequence. To translate this evolution into effective generative constraints, IMPACT derives sparse trajectories and dense region-level motion priors from the same physical rollout. Sparse trajectories constrain global object displacement, while dense priors guide diffusion sampling to encourage coherent object-region motion and suppress unwanted background changes. Their shared physical basis coordinates global and local motion guidance to reduce prediction--generation mismatch. Experiments on 100 controlled sphere--plane collision scenes spanning different material combinations and visual environments demonstrate improved physical plausibility over representative general-purpose, physics-grounded, and motion-controllable baselines. The project repository is available at \protect\url{https://weiweiwiewei.github.io/impact}.
\end{abstract}

%% file: Sections/1-introduction.tex
\section{Introduction}
\label{sec:introduction}

Generating physically consistent videos is essential for visual generative models to capture how the world evolves, with applications in embodied intelligence, robotics, and interactive simulation. \textit{Despite substantial advances in visual quality, generated videos still exhibit implausible contact responses, rebounds without energy dissipation, or abrupt motion termination.} These failures reveal a limitation beyond visual realism: individually plausible actions do not necessarily form a physically coherent interaction. 



\begin{figure}
    \centering
        \begin{overpic}[width=0.98\linewidth]{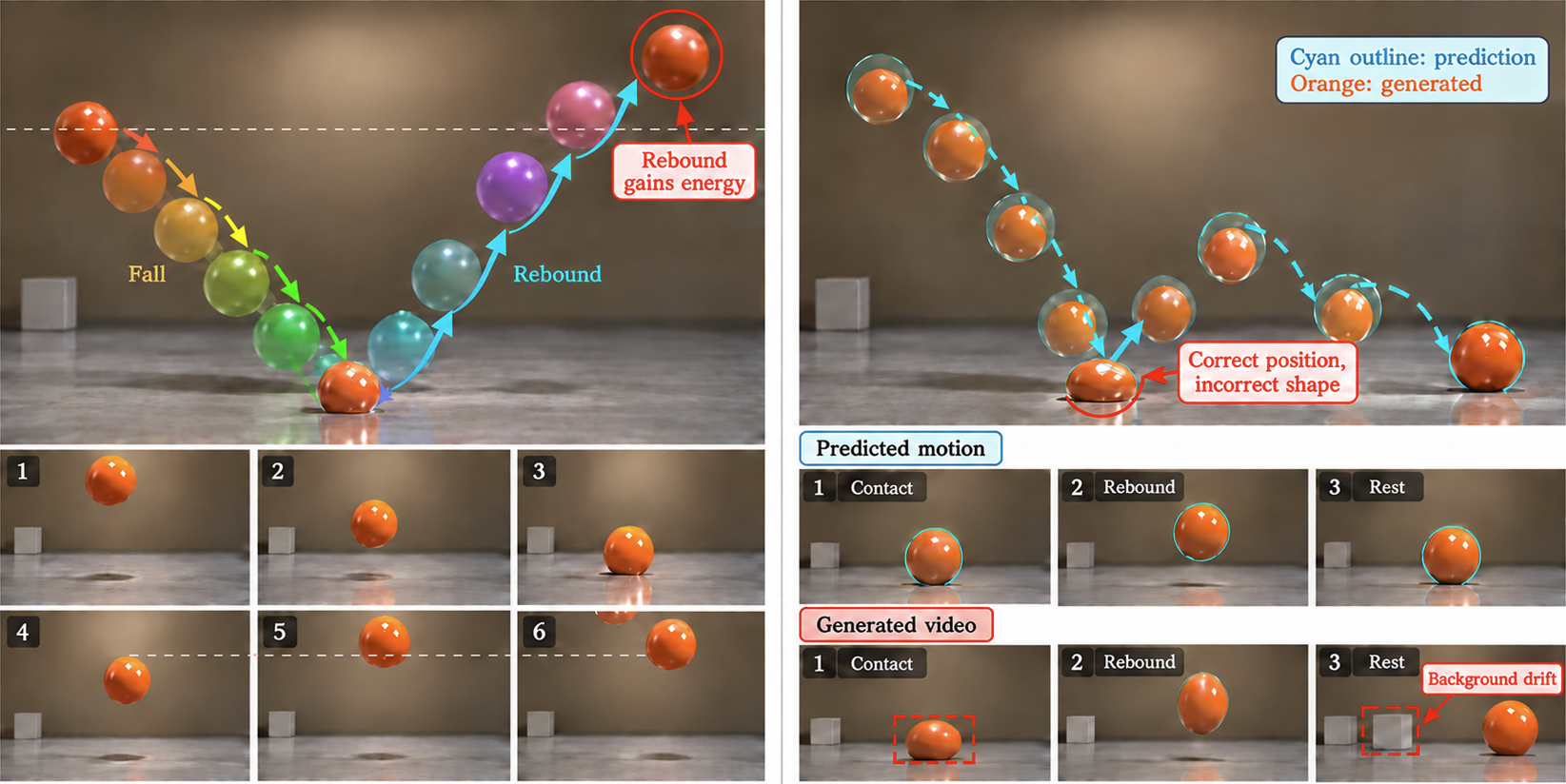}
        \put(2,-4){\footnotesize (a) Inconsistent motion evolution}
    \put(52,-4){\footnotesize (b) Prediction-generation mismatch }
    \end{overpic}
    \vspace{0.2cm}
    \caption{Illustration of two challenges in physically consistent video generation. (a) \textbf{Inconsistent motion evolution:} a sphere rebounds above its initial release height despite being released from rest, indicating an implausible gain in energy.
(b) \textbf{Prediction-generation mismatch:} even when object positions follow a physically valid prediction (cyan outlines), the generated video can exhibit object deformation, unstable contact, and background drift.}
    \label{fig:teaser}
\end{figure}

Existing approaches incorporate physical knowledge into video generation through language reasoning, physics-based simulation, or explicit motion control. Language-based methods, such as Physis-Lang~\cite{lu2026physis}, describe physical interactions and their consequences, but such descriptions do not explicitly determine impact timing, collision velocities, or energy dissipation. Physics-grounded approaches, including PhysGen~\cite{physgen} and MotionCraft~\cite{motioncraft}, introduce simulated dynamics to provide more concrete motion references, while motion-controllable methods guide synthesis through trajectories, optical flow, or spatial correspondences. Despite these advances, physically coherent motion specification and its faithful visual realization remain distinct requirements. A motion reference may describe a valid physical process without fully constraining how that process unfolds in the generated video. This distinction exposes a fundamental gap between physical-state prediction and generative motion realization.

Bridging this gap involves two challenges. The first is \textbf{inconsistent motion evolution}, where individually plausible actions fail to form a physically coherent sequence. As illustrated in Fig.~\ref{fig:teaser}(a), a sphere may rebound above its initial release height or continue bouncing without sufficient energy loss. The underlying difficulty lies in maintaining state-dependent consistency across successive physical events. Free-flight dynamics determine the incoming collision state, the collision response determines the outgoing velocity, and the remaining energy governs subsequent rebounds and settling. Without explicitly connecting these transitions, a generated sequence may depict recognizable physical events while violating the causal relationships among them. The second challenge is \textbf{prediction-generation mismatch}. Even when a physically valid motion prediction is available, its visual realization remains underdetermined. A center trajectory specifies the global displacement of an object but does not fully constrain how its entire region should move, how its boundary should interact with the supporting surface, or which background regions should remain stationary. Consequently, matching object-center positions does not necessarily guarantee consistent region-level motion, stable contact, or background preservation, as shown in Fig.~\ref{fig:teaser}(b).

To address these challenges, we propose \textbf{IMPACT}, a physics-state-driven image-to-video framework that couples explicit dynamics prediction with sparse-dense physical guidance. IMPACT predicts a connected sequence of physical states and uses that sequence to guide both global object displacement and region-level motion during synthesis. 
We study sphere-plane interactions involving free fall, successive collisions, and eventual settling to evaluate the physical consistency of predicted motion and its visual realization across different materials and environments.

To address inconsistent motion evolution, IMPACT makes each stage of the interaction depend explicitly on the physical state reached in the preceding stage. The Image-to-Physics (I2P) module establishes the conditions governing this evolution by estimating a compact, motion-relevant physical configuration from the initial image and text prompt. Building on this configuration, the Physics State Generator (PSG) evolves position and velocity through continuous free-flight dynamics until contact triggers a discrete collision response. The updated velocity then governs the next rebound, linking successive motion stages through a common dynamical process. Learnably calibrated restitution controls the collision response, while explicit rebound attenuation and settling model how repeated impacts progressively bring the sphere to rest. This connected physical evolution is projected into sparse image-space trajectories, allowing a pretrained trajectory-control interface to guide global object displacement according to the predicted dynamics.

To reduce prediction-generation mismatch, IMPACT complements these sparse trajectories with dense physical guidance during diffusion sampling. Rather than relying on center positions alone, we derive dense object masks and optical flow from the same PSG rollout to provide region-level motion priors. Through energy-based latent guidance, these priors encourage motion across the generated object region to follow the physical prediction while suppressing unwanted background changes. Sparse trajectories anchor the overall motion path, whereas dense priors guide how that motion is realized across the object and surrounding scene. Their shared physical origin ensures that both forms of guidance refer to the same predicted interaction, providing complementary constraints on its visual realization.

\begin{figure*}[!t]
\centering
\includegraphics[width=0.98\textwidth]{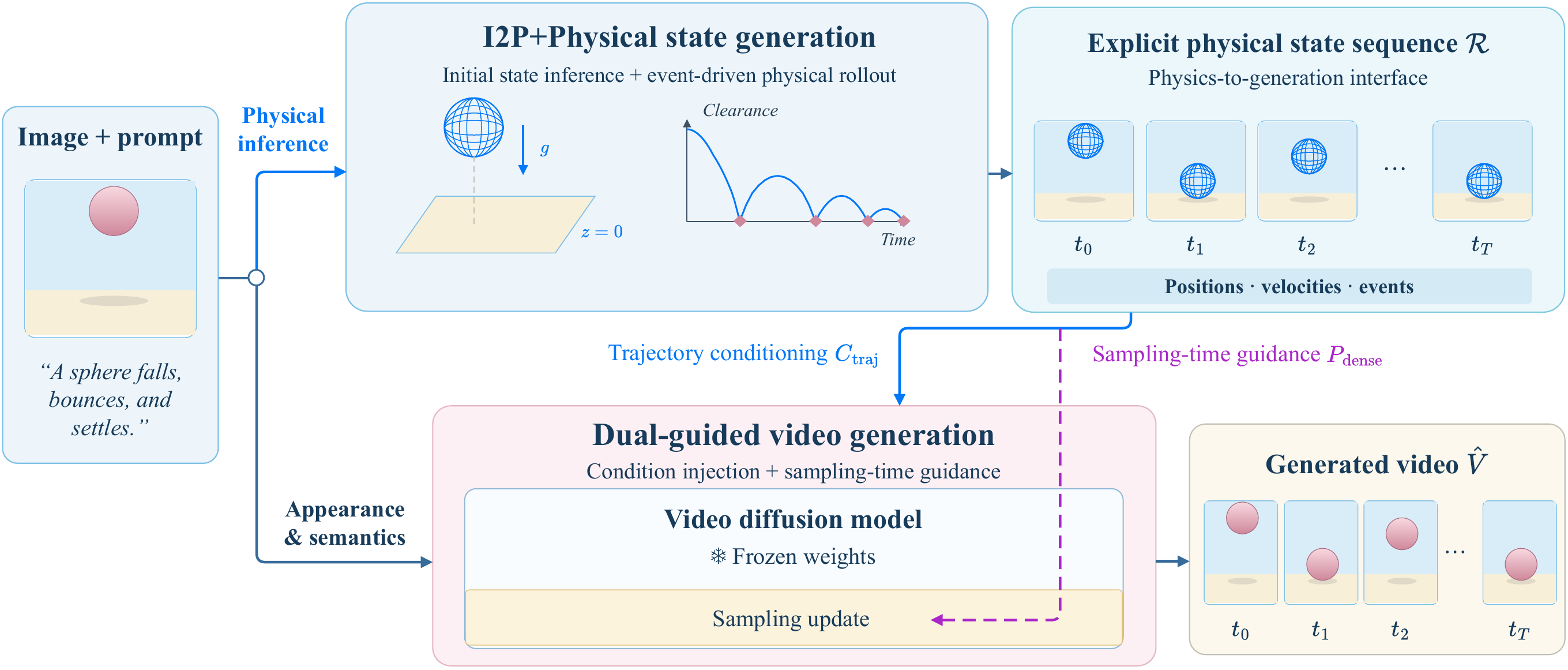}
\caption{Overview of IMPACT.
The Image-to-Physics (I2P) module estimates a compact physical configuration from an initial image and a text prompt.
The Physics State Generator (PSG) combines continuous free-flight dynamics with discrete collision events to model free fall, rebound attenuation, and settling.
Sparse trajectories and dense region-level motion priors are derived from the same physical rollout and provide complementary guidance for video synthesis.
Both the video backbone and the pretrained trajectory-control module remain frozen.
}
\label{fig:framework}
\end{figure*}

We evaluate IMPACT on 100 controlled sphere-plane collision scenes with different material combinations and visual environments. Human and vision-language-model evaluations assess physical event completeness, free-fall dynamics, contact realism, rebound attenuation, and overall physical plausibility. The results show improved physical plausibility over representative general-purpose, physics-grounded, and motion-controllable baselines. 

Our main contributions are summarized as follows:
\begin{itemize}
\item We propose \textbf{IMPACT}, an image-to-video framework to extend physical consistency beyond motion prediction to two interconnected requirements: consistency in physical-state evolution and faithful visual realization of predicted motion.

\item We develop an event-driven hybrid dynamics model that unifies continuous motion and discrete collision events, explicitly establishing causal dependencies across successive physical events.

\item We translate a unified physical-state evolution into complementary sparse--dense generative constraints, bridging the gap between physical motion prediction and video generation.
\end{itemize}

The remainder of this paper is organized as follows. Section~\ref{sec:related_work} reviews related work. Section~\ref{sec:method} presents the proposed method. Section~\ref{sec:experiment} provides extensive experiments and analysis. Finally, Section~\ref{sec:conclusion} concludes this paper.

%% file: Sections/2-relatedwork.tex
\section{Related Work}
\label{sec:related_work}

\subsection{Physics-Grounded Video Generation}

Physics-grounded video generation seeks to bridge the gap between visual realism and physically plausible motion. One line of research explicitly couples physical simulation with visual synthesis. MotionCraft employs simulation-derived optical flow to warp diffusion noise, enabling physically guided motion generation. PhysGen performs image-space rigid-body simulation and uses generative refinement to synthesize realistic videos, while PhysGen3D~\cite{physgen3d} reconstructs an interactive 3D scene from a single image to simulate object interactions. DreamPhysics~\cite{dreamphysics} estimates material properties to model deformable object dynamics. More recently, PSIVG~\cite{psivg} reconstructs object geometry and 4D scene information from a generated template video, performs physical simulation, and uses the resulting motion cues to guide video synthesis, with additional test-time optimization for texture consistency. These approaches demonstrate the effectiveness of simulation-based motion guidance, while differing in their requirements for scene reconstruction, physical initialization, and integration with generative models.

Another line of research incorporates physical knowledge through learned dynamics, latent representations, or physics-aware optimization. NewtonGen~\cite{newtongen} introduces trainable Neural Newtonian Dynamics to predict physical motion and provide dynamical guidance for video synthesis. PhysVideoGenerator~\cite{physvideogenerator} incorporates learned latent physics representations into diffusion transformers, while other approaches explore physical priors, language-guided reasoning, and learned state evolution~\cite{latentphysics,vlipp,diffphy,physctrl,newton}. More recently, PhyCo~\cite{phyco} learns controllable physical responses through property-conditioned fine-tuning and reward optimization, whereas PhysRVG~\cite{physrvg} employs physics-aware reinforcement learning to improve rigid-body motion generation. These methods introduce physical knowledge at different stages of the generation process, offering complementary alternatives to explicit simulation.

Despite these advances, rigid-body impacts remain challenging because continuous gravitational motion is interrupted by discrete contact events that alter object velocity and dissipate energy. Accurately modeling repeated rebounds therefore requires coordinated treatment of free-flight dynamics, collision responses, and eventual settling. This motivates an event-driven formulation that explicitly resolves physical state transitions throughout the motion process.

\subsection{Motion-Controllable Video Generation}

Motion-controllable video generation regulates object or camera motion through explicit spatial and temporal conditions. Trajectory-based methods provide intuitive control over object displacement and motion paths. TrackGo~\cite{trackgo}, Motion Prompting~\cite{motionprompting}, and MotionPro~\cite{motionpro} introduce trajectory- or region-level conditions to guide video synthesis. Tora~\cite{tora} further encodes trajectories into hierarchical spatiotemporal motion representations and integrates them into diffusion transformers, while Wan-Move~\cite{wanmove} propagates first-frame features along prescribed motion paths. These approaches demonstrate the effectiveness of trajectory guidance for specifying global motion structures and controlling object movement.

Beyond sparse trajectories, dense motion representations provide finer spatial constraints for video generation. Motion-I2V~\cite{motioni2v} separates motion-field prediction from video synthesis, allowing explicit motion representations to guide temporal generation. Go-with-the-Flow~\cite{gowiththeflow} introduces optical-flow-based noise warping to control motion, while FloVD~\cite{flovd} employs optical-flow representations for controllable video synthesis. MagicMotion~\cite{magicmotion} further supports multiple levels of trajectory conditions, including masks, bounding boxes, and sparse boxes, enabling motion control at different spatial granularities. Although these representations offer complementary levels of control, their ability to produce physically plausible motion depends on the underlying motion signals. In particular, specifying object positions or displacement fields does not inherently enforce gravitational acceleration, collision responses, or energy dissipation.

\subsection{Training-Free Guidance for Video Diffusion}

Training-free guidance provides a flexible mechanism for controlling pretrained diffusion models without updating their parameters. Diffusion Posterior Sampling~\cite{chung2023dps}, FreeDoM~\cite{freedom}, and TFG~\cite{tfg} introduce inference-time guidance strategies based on likelihoods, differentiable objectives, or energy functions. By evaluating guidance objectives on denoising predictions and propagating gradients to the sampling variables, these methods steer generation toward desired constraints while preserving the pretrained model. Such approaches provide a general foundation for incorporating external information into diffusion sampling without task-specific retraining.

For video generation, training-free control has been extended to motion manipulation and temporal consistency. Motion-Zero~\cite{motionzero} combines position-aware noise priors with spatial and temporal attention constraints to control object trajectories. Related studies investigate motion representations in pretrained video models~\cite{moft}, while MotionFlow~\cite{motionflow} performs latent optimization using attention-derived motion cues. DiTFlow~\cite{ditflow} extracts motion information from diffusion-transformer attention and employs optimization-based guidance for motion transfer. These approaches demonstrate that motion can be influenced through noise, attention, or latent-space optimization without updating the generative backbone. However, enforcing physical dynamics additionally requires temporally coordinated constraints that reflect the evolution of object states, particularly across discontinuous collision events.

%% file: Sections/3-method.tex
\section{Method}
\label{sec:method}

\subsection{Problem Formulation and Framework Overview}
\label{sec:method_overview}

Given an initial image $I_0\in\mathbb R^{H\times W\times3}$
and a text condition $y$, the objective is to generate a video
$\widehat{\mathcal V}=\{\widehat I_t\}_{t=0}^{T-1}$
that preserves the appearance and scene structure of $I_0$
while exhibiting physically consistent motion.
The considered scenario consists of a single rigid sphere
released from rest above a stationary rigid plane, observed
by a fixed, approximately side-view camera.
The target motion comprises gravitational free fall,
ground impact, successive rebounds, and eventual settling.

The proposed framework decouples physical motion modeling
from video synthesis. As illustrated in
Fig.~\ref{fig:framework}, an Image-to-Physics (I2P)
module first estimates the initial physical state and camera
parameters. An event-driven Physics State Generator (PSG)
then propagates the state through continuous free flight
and discrete collision events, producing a structured
physical rollout. Finally, the rollout provides both
sparse trajectory conditions and dense region-level
constraints for a pretrained video diffusion model.
The video backbone and trajectory-control module remain
frozen, while only the physical parameter calibration
components are trained.

\subsection{Initial Physical State Estimation}
\label{sec:method_i2p}

\begin{figure}[t]
    \centering
    \includegraphics[width=\columnwidth]{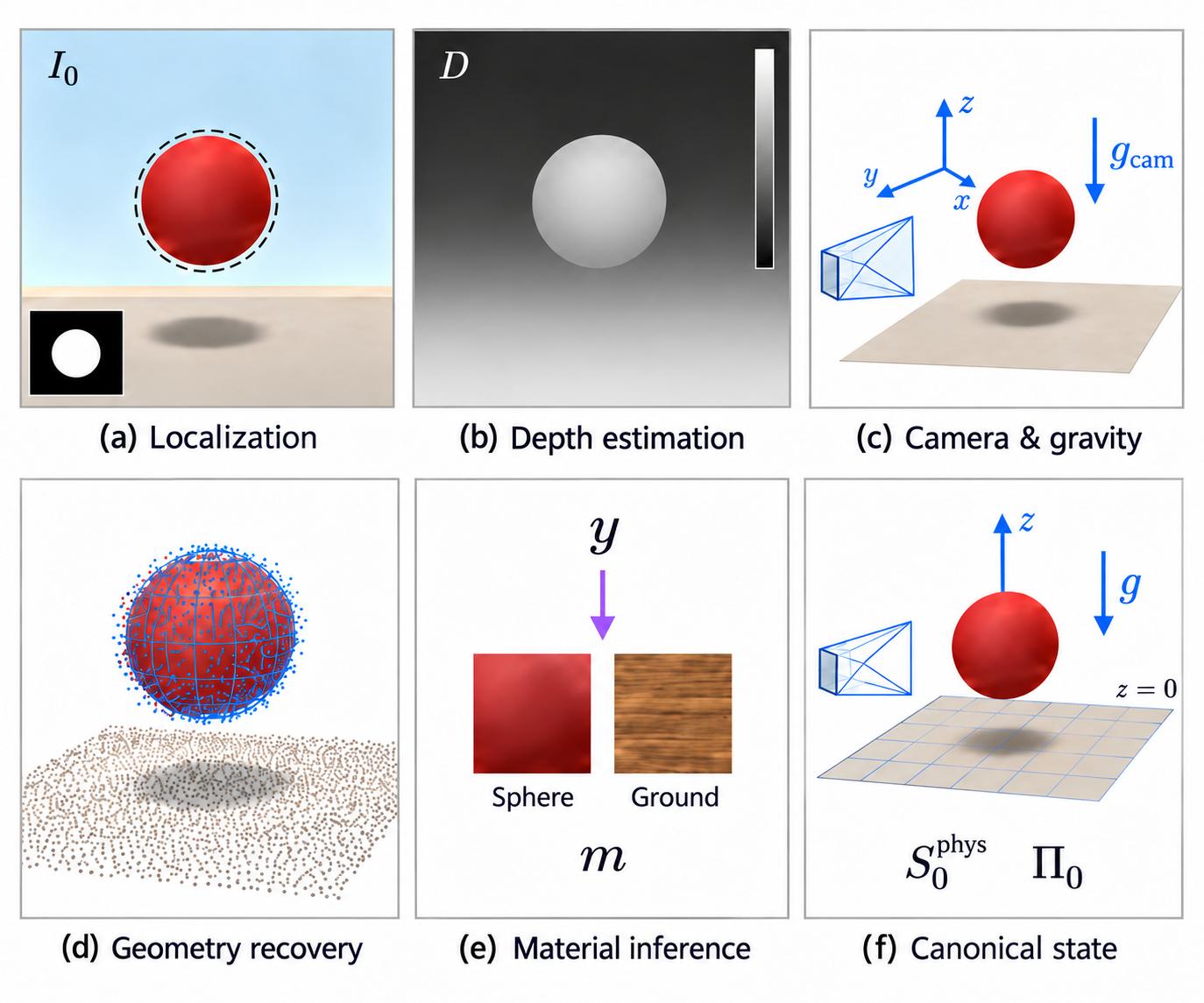}
    \vspace{-0.8cm}
    \caption{Visualization of the Image-to-Physics (I2P).
    (a) Object localization and segmentation.
    (b) Monocular depth estimation.
    (c) Camera calibration and gravity estimation.
    (d) Sphere--plane geometry recovery from
    back-projected observations.
    (e) Material conditioning from the text prompt.
    (f) Construction of the plane-aligned canonical
    physical state and camera parameters.
    }
    \label{fig:i2p}
\end{figure}

I2P extracts the physical quantities required for
sphere--plane dynamics without reconstructing a complete
three-dimensional scene. The initial physical state is
represented as
\begin{equation}
\mathcal S_0^{\mathrm{phys}}
=
\{\mathbf P_0,\mathbf V_0,r,\mathbf n,c,m\},
\label{eq:initial_physical_state}
\end{equation}
where $\mathbf P_0,\mathbf V_0\in\mathbb R^3$ denote
the initial sphere center and velocity, $r$ is the sphere
radius, $(\mathbf n,c)$ specifies the supporting plane
with $\|\mathbf n\|_2=1$, and $m$ denotes the
sphere--ground material condition.
The initial velocity is set to $\mathbf V_0=\mathbf0$.
Camera parameters
$\Pi_0=\{\mathbf K,\mathbf R_{cw},\mathbf t_{cw}\}$
are estimated separately for subsequent projection.

As illustrated in Fig.~\ref{fig:i2p}, I2P
combines object localization, depth estimation,
and camera calibration to recover the
motion-relevant sphere--plane geometry,
while the material condition is obtained
from the text prompt.
The resulting geometric configuration is
transformed into a plane-aligned canonical
coordinate system, where the supporting plane
lies at $z=0$ and gravity acts along the
negative $z$-axis.
For a sphere center $\mathbf P_t$, the signed contact gap is
\begin{equation}
\delta_t
=
\mathbf n^\top\mathbf P_t+c-r.
\label{eq:contact_gap}
\end{equation}
A positive gap indicates separation, while zero and negative
values correspond to contact and penetration, respectively.
This quantity provides the geometric condition for
subsequent event-driven state evolution.

\subsection{Event-Driven Physics State Generator}
\label{sec:method_psg}

Sphere--plane motion exhibits hybrid dynamics, where
continuous gravitational motion is interrupted by discrete
collision events. PSG models these two forms of state
evolution explicitly, rather than directly regressing
future trajectories. Given the initial physical state,
it generates a structured rollout
\begin{equation}
\mathcal R
=
\{\mathbf P_t,\mathbf V_t,\chi_t,e_t\}_{t=0}^{T-1},
\label{eq:physical_rollout}
\end{equation}
where $\mathbf P_t$ and $\mathbf V_t$ denote the sphere
position and velocity, $\chi_t\in\{0,1\}$ indicates
whether a collision occurs within the corresponding
time interval, and $e_t$ is the effective restitution
coefficient associated with an impact.

\subsubsection{Physical Parameter Calibration}
\label{sec:psg_calibration}

PSG employs lightweight parameter calibration to account
for uncertainty in the physical quantities recovered from
a single image. Rather than predicting object trajectories,
the calibration module adjusts the parameters governing
their evolution. Specifically, the nominal gravitational
magnitude $g_0$ is corrected through a bounded function:
\begin{equation}
\begin{aligned}
\Delta g
&=
\epsilon_g g_0
\tanh\left(f_\phi(\mathcal S_0^{\mathrm{phys}})\right),\\
g_{\mathrm{eff}}
&=
g_0+\Delta g,
\qquad
\mathbf g=-g_{\mathrm{eff}}\mathbf n,
\end{aligned}
\label{eq:gravity_calibration}
\end{equation}
where $f_\phi$ is a learnable calibration network and
$\epsilon_g$ limits the relative correction.
The calibrated gravitational magnitude compensates
for scale uncertainty in monocular scene estimation
without introducing unrestricted changes to the
underlying dynamics.

Collision response is parameterized separately using
a material- and velocity-dependent restitution coefficient:
\begin{equation}
\begin{aligned}
e_m(v)
={}&e_{\min,m}
+(e_{\max,m}-e_{\min,m})\\
&\cdot
\left(1+\frac{v}{v_{\mathrm{ref}}}\right)^{-k_m},
\end{aligned}
\label{eq:restitution_curve}
\end{equation}
where $e_{\min,m}$, $e_{\max,m}$, and $k_m$
control the restitution behavior of material condition
$m$, subject to
$0\le e_{\min,m}\le e_{\max,m}\le1$
and $k_m\ge0$.
These parameterizations allow physical properties
to be calibrated while retaining explicit state evolution.

\subsubsection{Event-Driven State Evolution and Collision Response}
\label{sec:psg_dynamics}

Between collision events, the sphere follows
constant-acceleration motion. Given the state
$(\mathbf P_t,\mathbf V_t)$, its continuous evolution
over an interval $\tau$ is
\begin{equation}
\begin{aligned}
\mathbf P_t(\tau)
&=
\mathbf P_t+\mathbf V_t\tau
+\frac{1}{2}\mathbf g\tau^2,\\
\mathbf V_t(\tau)
&=
\mathbf V_t+\mathbf g\tau.
\end{aligned}
\label{eq:free_flight}
\end{equation}
Rather than applying collision updates only at discrete
frame boundaries, PSG resolves the first valid contact
time within each simulation interval $\Delta t$:
\begin{equation}
\tau^*
=
\min
\left\{
\tau\in(0,\Delta t]
\mid
\delta_t(\tau)=0
\right\},
\label{eq:time_of_impact}
\end{equation}
subject to the sphere approaching the supporting plane.
The corresponding pre-impact velocity is
$\mathbf V^-=\mathbf V_t(\tau^*)$.

At contact, the normal component of velocity is updated
according to the effective restitution coefficient
$e_t=e_m(-\mathbf n^\top\mathbf V^-)$:
\begin{equation}
\mathbf V^+
=
\mathbf V^-
-
(1+e_t)
(\mathbf n^\top\mathbf V^-)\mathbf n.
\label{eq:collision_response}
\end{equation}
The remaining portion of the current time interval
is then propagated from the post-impact state.
By explicitly locating impact events and resolving
the corresponding velocity transitions, PSG maintains
the temporal ordering of free flight and collision
while reducing contact-time errors associated with
frame-boundary updates.

Successive rebounds are obtained by repeatedly applying
the continuous evolution and collision response.
When the predicted rebound height becomes sufficiently
small, a short settling transition brings the sphere
to a nonpenetrating rest state, avoiding unresolved
micro-bounces and abrupt termination.
Numerical contact correction and settling thresholds
are detailed in the supplementary material.

\subsubsection{Physical-State Supervision}
\label{sec:psg_training}

The learnable parameters of PSG are optimized using
two complementary physical-state objectives.
The position loss constrains the complete state
trajectory, whereas the impact loss specifically
supervises the discontinuous velocity changes
induced by collisions. The training objective is
\begin{equation}
\mathcal L_{\mathrm{PSG}}
=
\mathcal L_{\mathrm{pos}}
+
\lambda_{\mathrm{impact}}\mathcal L_{\mathrm{impact}},
\label{eq:psg_loss}
\end{equation}
where $\lambda_{\mathrm{impact}}$ is the hyperparameter and is set to 0.2 in the experiments. 

Given predicted sphere-center positions
$\widehat{\mathbf P}_{b,t}$ and reference positions
$\mathbf P_{b,t}$, the position loss is defined as
\begin{equation}
\mathcal L_{\mathrm{pos}}
=
\frac{1}{3BT}
\sum_{b=1}^{B}
\sum_{t=0}^{T-1}
\sum_{d=1}^{3}
\rho\left(
\widehat P_{b,t,d}-P_{b,t,d}
\right),
\label{eq:position_loss}
\end{equation}
where $B$ is the batch size, $T$ is the sequence length,
and $\rho(\cdot)$ denotes the Smooth-$L_1$ function.
This term provides trajectory-level supervision across
free-flight, rebound, and settling phases.

To constrain collision response, the impact loss measures
the discrepancy between predicted and reference
post-impact normal velocities:
\begin{equation}
\mathcal L_{\mathrm{impact}}
=
\frac{1}{|\mathcal I|}
\sum_{(b,t)\in\mathcal I}
\rho\left(
\mathbf n_b^\top
\widehat{\mathbf V}_{b,t}^{+}
-
\mathbf n_b^\top
\mathbf V_{b,t}^{+}
\right),
\label{eq:impact_loss}
\end{equation}
where $\mathcal I$ denotes the set of annotated
impact states. The two objectives jointly constrain
continuous motion and discrete collision response,
while the resulting trajectories remain determined
by explicit physical dynamics rather than direct
neural prediction. 
\begin{figure}[!t]
    \centering
    \includegraphics[width=\columnwidth]
    {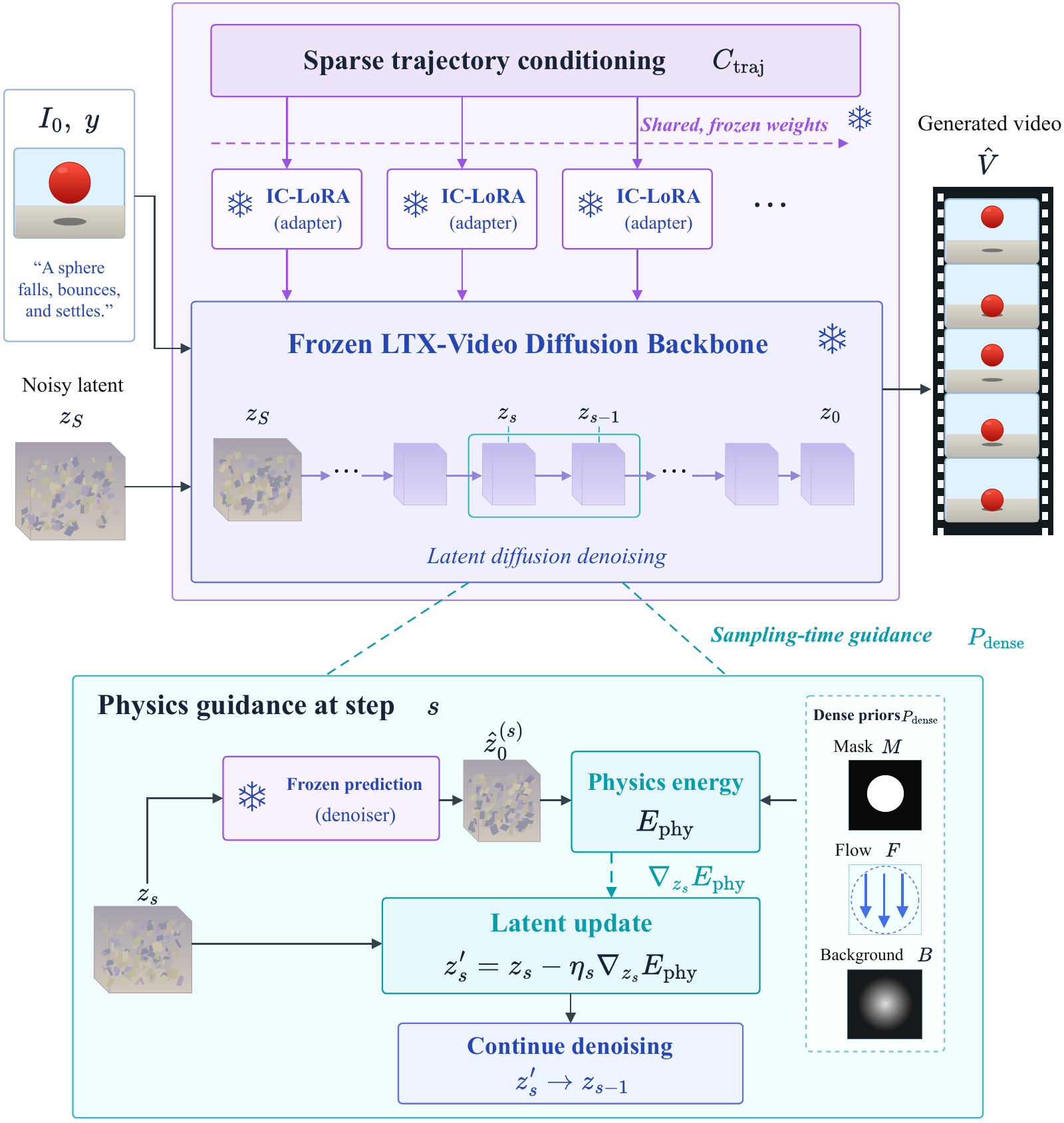}
    \caption{Complementary Sparse-Dense Physical Guidance.
    The physics-derived sparse trajectory
    $\mathcal C_{\mathrm{traj}}$ conditions
    the frozen LTX-Video backbone through
    pretrained Motion Track IC-LoRA.
    Dense physical priors
    $\mathcal P_{\mathrm{dense}}$ define
    object-motion and background-stability
    constraints on the predicted clean latent.
    Their gradients update the noisy latent
    during selected denoising steps without
    modifying the pretrained model parameters.
    }
    \label{fig:dual_guided}
\end{figure}

\subsection{Physics-Coherent Video Generation}
\label{sec:method_video}

The physical rollout specifies the target motion,
but its consistency is not automatically preserved
during video generation. As illustrated in
Fig.~\ref{fig:dual_guided}, we employ two
complementary guidance mechanisms within a frozen
video diffusion backbone. Sparse trajectory
conditioning transfers the predicted global motion
and event timing through pretrained Motion Track
IC-LoRA, while dense physical priors impose
region-level constraints through sampling-time
latent guidance. Both forms of guidance originate
from the same physical rollout, connecting global
motion control with local motion consistency
and background stability.

\subsubsection{Physics-Derived Motion Conditioning}
\label{sec:motion_conditioning}

The three-dimensional sphere centers are projected
into the image plane as
$\mathbf p_t=\pi(\mathbf P_t;\Pi_0)$,
with visibility indicators $q_t\in\{0,1\}$.
The projected trajectory is converted into the
conditioning format of the pretrained Motion Track
IC-LoRA:
\begin{equation}
\mathcal C_{\mathrm{traj}}
=
\mathcal H_{\mathrm{traj}}
\left(
\{\mathbf p_t,q_t\}_{t=0}^{T-1}
\right).
\label{eq:trajectory_condition}
\end{equation}
This sparse condition transfers the physically
determined global motion and event timing to the
video generator without retraining the trajectory
control module.

To provide region-level constraints, the same
trajectory is used to construct target masks $M_t$
and a cumulative displacement field referenced
to the initial frame:
\begin{equation}
\mathbf F_t(\mathbf x)
=
M_0(\mathbf x)
(\mathbf p_t-\mathbf p_0),
\label{eq:physics_displacement}
\end{equation}
where $M_0$ is the initial object mask.
Under the fixed-camera and approximately
constant-depth setting, this field describes the
dominant translational motion of the target.
Reliable static background regions are defined by
\begin{equation}
B_t
=
1-\operatorname{Dilate}(M_0\lor M_t),
\label{eq:background_mask}
\end{equation}
excluding the initial and current object regions
and their uncertain boundaries.

The resulting dense prior
$\mathcal P_{\mathrm{dense}}
=\{M_t,\mathbf F_t,B_t,q_t\}_{t=0}^{T-1}$
is aligned with the spatiotemporal resolution of
the video VAE:
\begin{equation}
\begin{aligned}
\mathcal P_{\mathrm{dense}}^z
&=
\mathcal A_z(\mathcal P_{\mathrm{dense}})\\
&=
\{M_\ell^z,\mathbf F_\ell^z,
B_\ell^z,q_\ell^z\}_{\ell=0}^{T_z-1}.
\end{aligned}
\label{eq:latent_physics_prior}
\end{equation}
Here $\mathcal A_z$ denotes spatial downsampling
and temporal alignment, and $T_z$ is the latent
sequence length. Unlike sparse trajectory conditions,
these dense priors are not directly injected as
network inputs; instead, they define constraints
for the subsequent denoising process.

\subsubsection{Sampling-Time Physics-Guided Diffusion}
\label{sec:physics_guidance}

As illustrated in the lower part of
Fig.~\ref{fig:dual_guided}, dense physical
priors complement sparse trajectory conditioning
by imposing differentiable constraints on
object-region motion and background stability
during diffusion sampling.
Therefore, dense physical priors are incorporated
through differentiable latent-space constraints
during diffusion sampling.

Let $\mathbf z_s^v$ denote the noisy video latent
at denoising step $s$, with noise level $\sigma_s$.
Conditioned on the initial image, text, and
physics-derived trajectory, the frozen denoiser
produces an estimate of the clean latent:
\begin{equation}
\widehat{\mathbf z}_0^{(s)}
=
\mathcal D_\theta
\left(
\mathbf z_s^v,\sigma_s;
I_0,y,\mathcal C_{\mathrm{traj}}
\right).
\label{eq:clean_latent_prediction}
\end{equation}
The initial target representation is transported
to its physically predicted location using
the cumulative displacement field:
\begin{equation}
\widetilde{\mathbf z}_{0,\ell}^{(s)}
=
\mathcal W
\left(
\widehat{\mathbf z}_{0,0}^{(s)},
\mathbf F_\ell^z
\right),
\label{eq:latent_warp}
\end{equation}
where $\mathcal W$ denotes differentiable spatial
warping.

The object-motion consistency energy is defined as
\begin{equation}
E_{\mathrm{obj}}^{(s)}
=
\frac{
\displaystyle\sum_{\ell=1}^{T_z-1}
q_\ell^z
\left\|
M_\ell^z\odot
\left(
\widehat{\mathbf z}_{0,\ell}^{(s)}
-
\widetilde{\mathbf z}_{0,\ell}^{(s)}
\right)
\right\|_1
}{
\displaystyle\sum_{\ell=1}^{T_z-1}
q_\ell^z\|M_\ell^z\|_1+\varepsilon
},
\label{eq:object_energy}
\end{equation}
where $\odot$ denotes elementwise multiplication.
This term constrains the generated target region
to remain consistent with the physics-derived
displacement, rather than relying solely on
point-level trajectory alignment.

A complementary background-stability energy
penalizes changes in regions expected to remain static:
\begin{equation}
E_{\mathrm{bg}}^{(s)}
=
\frac{
\displaystyle\sum_{\ell=1}^{T_z-1}
q_\ell^z
\left\|
B_\ell^z\odot
\left(
\widehat{\mathbf z}_{0,\ell}^{(s)}
-
\widehat{\mathbf z}_{0,0}^{(s)}
\right)
\right\|_1
}{
\displaystyle\sum_{\ell=1}^{T_z-1}
q_\ell^z\|B_\ell^z\|_1+\varepsilon
},
\label{eq:background_energy}
\end{equation}
where $\varepsilon>0$ provides numerical stability.

Following differentiable sampling-time guidance
\cite{chung2023dps,bansal2024universal},
the two energies are combined as
\begin{equation}
E_{\mathrm{phy}}^{(s)}
=
\lambda_{\mathrm{obj}}^{(s)}
E_{\mathrm{obj}}^{(s)}
+
\lambda_{\mathrm{bg}}^{(s)}
E_{\mathrm{bg}}^{(s)},
\label{eq:physics_energy}
\end{equation}
and the current noisy latent is updated by
\begin{equation}
\widetilde{\mathbf z}_s^v
=
\mathbf z_s^v
-
\eta_s
\nabla_{\mathbf z_s^v}
E_{\mathrm{phy}}^{(s)}.
\label{eq:physics_guidance}
\end{equation}
Although the energy is evaluated on the predicted
clean latent, its gradient is propagated to the
current noisy latent through the frozen denoising
computation. The sampler subsequently continues
from $\widetilde{\mathbf z}_s^v$ without updating
the video backbone or trajectory-control parameters.

Physics guidance is applied during an intermediate
portion of the denoising process, with scheduled
guidance strength and energy weights to limit
interference with visual synthesis.
The physical rollout and corresponding priors are
computed once before sampling, without repeated
physical simulation or video-level motion estimation
inside the denoising loop. In this manner, the sparse
condition specifies the global motion, while the
dense latent constraints regulate region-level
consistency using the same physical rollout.

\begin{figure}[t]
    \centering
    \includegraphics[width=\columnwidth]{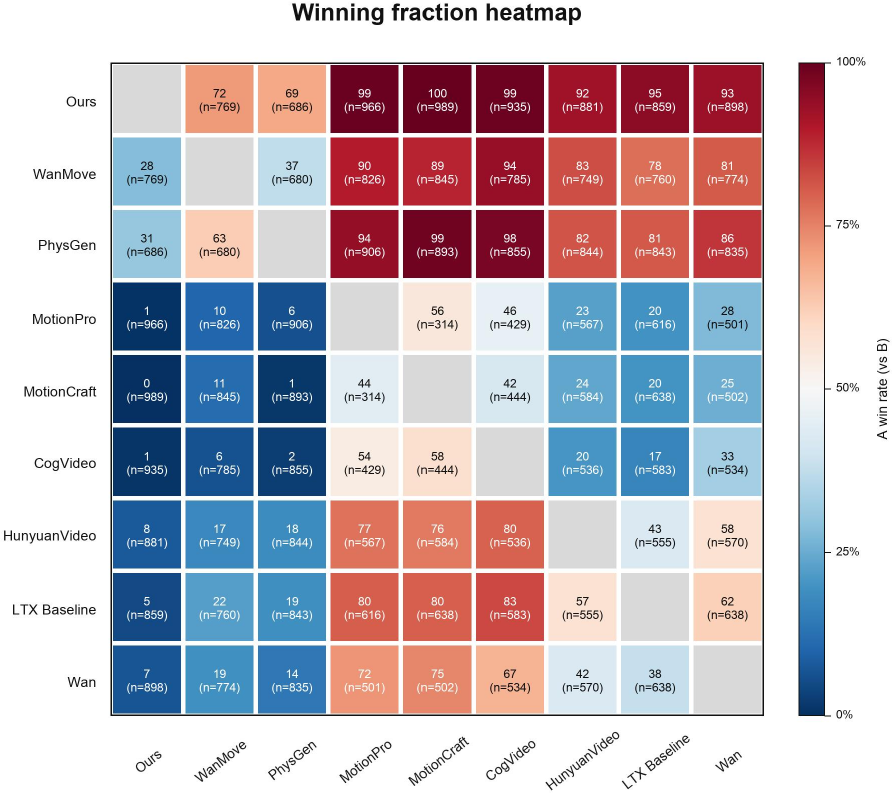}
    \caption{
    Pairwise human rating win-rate heatmap based on
    the Overall Preference (OPP) metric.
    Each cell reports the percentage of non-tied
    comparisons in which the row method receives
    a higher rating than the column method.
    Ratings are paired by test video and evaluator,
    with ties excluded from the calculation.
    Warmer colors indicate higher winning fractions.
    }
    \label{fig:opp_winrate}
\end{figure}

\begin{figure*}[t]
\centering
\IfFileExists{Figures/fram.pdf}{%
  \includegraphics[width=\textwidth]{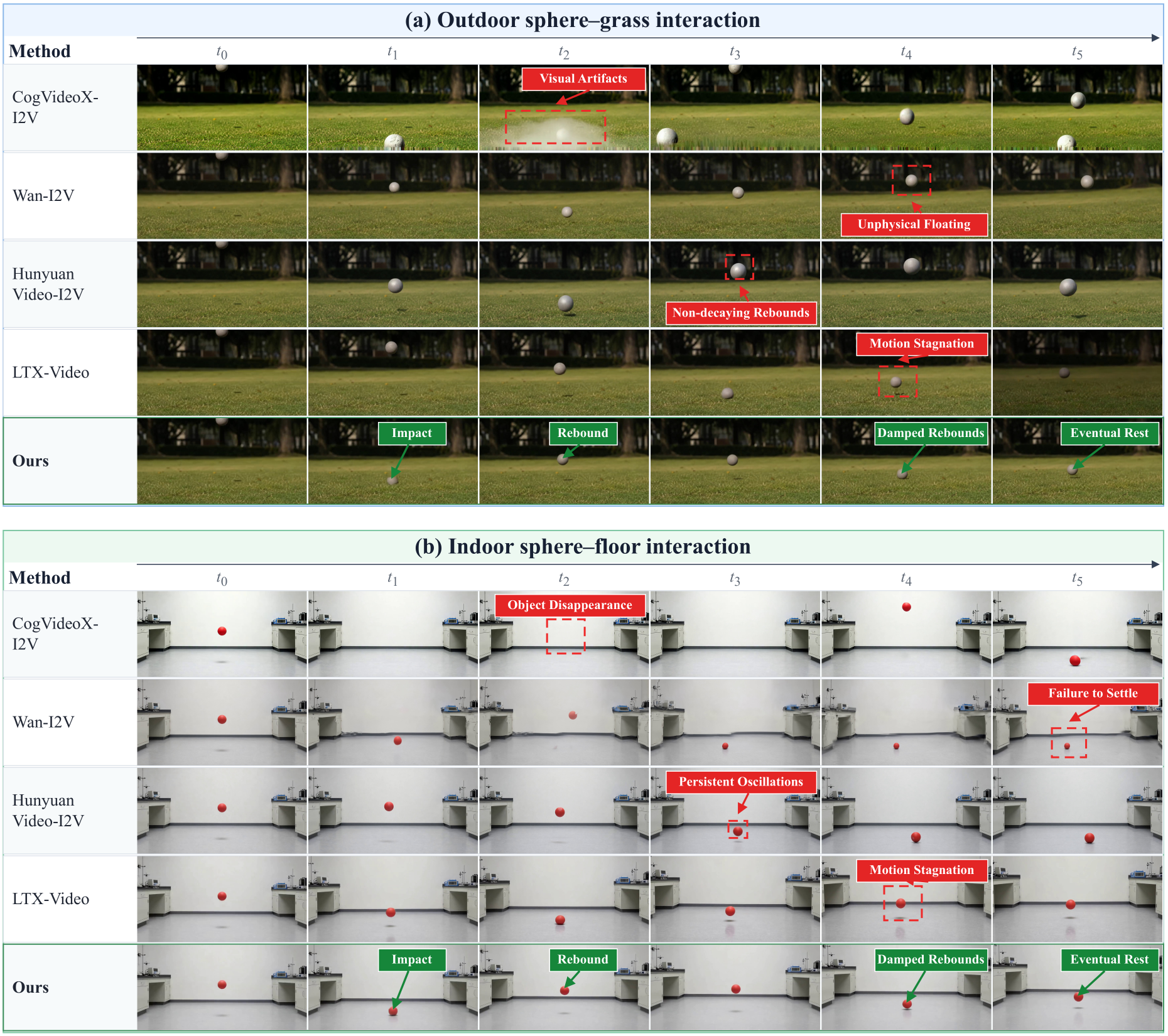}%
}{%
  \fbox{\parbox[c][1.1in][c]{0.91\textwidth}{\centering Original qualitative comparison figure: Figures/fram.pdf}}%
}
\vspace{-0.5cm}
\caption{Qualitative comparison of the evaluated free-fall and bouncing sequences. Representative frames illustrate free fall, contact, rebound attenuation, and settling.}
\label{fig:qualitative_comparison}
\end{figure*}


\begin{table}[t]
\centering
\caption{Inference configurations and additional input
requirements of the compared methods.}
\label{tab:inference_config}
\setlength{\tabcolsep}{4.5pt}
\renewcommand{\arraystretch}{1.12}
\begin{tabular}{@{}lcccc@{}}
\toprule
\textbf{Method} &
\textbf{Resolution} &
\textbf{Frames} &
\textbf{FPS} &
\textbf{Additional Input} \\
\midrule
CogVideoX-I2V    & $1024\times512$ & 81 & 25 & None \\
HunyuanVideo-I2V & $1344\times672$ & 81 & 25 & None \\
Wan-I2V          & $1024\times512$ & 81 & 25 & None \\
LTX-Video        & $1024\times512$ & 81 & 25 & None \\
\addlinespace[3pt]
PhysGen          & $512\times512$ & 81 & 25 &
Physical parameters \\
MotionCraft      & $512\times512$ & 81 & 25 &
Simulated motion \\
MotionPro        & $512\times512$ & 16 & 8 &
Trajectory + Mask \\
WanMove          & $880\times432$ & 81 & 25 &
Trajectory \\
\midrule
\textbf{IMPACT (Ours)} &
$1024\times512$ & 81 & 25 &
\textbf{None (automatic)} \\
\bottomrule
\end{tabular}
\end{table}

\begin{table*}[t]
\centering
\caption{
Quantitative comparison with general-purpose,
physics-grounded, and motion-controllable video
generation methods under both human and VLM evaluations.
Human evaluation scores are reported as the mean
$\pm$ standard deviation of evaluator-level mean
ratings on a 1-5 scale.
}
\label{tab:overall_comparison}

\footnotesize
\setlength{\tabcolsep}{4pt}
\renewcommand{\arraystretch}{1.12}

\resizebox{\textwidth}{!}{%
\begin{tabular}{l*{12}{c}}
\toprule

\multirow{2}{*}{\textbf{Method}}
& \multicolumn{6}{c}{\textbf{Human Evaluation}}
& \multicolumn{6}{c}{\textbf{VLM Evaluation}} \\

\cmidrule(lr){2-7}
\cmidrule(lr){8-13}

& PEC$\uparrow$
& FDR$\uparrow$
& CCR$\uparrow$
& REC$\uparrow$
& OPP$\uparrow$
& \mbox{CFR (\%)$\downarrow$}
& PEC$\uparrow$
& FDR$\uparrow$
& CCR$\uparrow$
& REC$\uparrow$
& OPP$\uparrow$
& \mbox{CFR (\%)$\downarrow$} \\

\midrule


\multicolumn{13}{l}{
\textit{General-Purpose Video Generators}
} \\

CogVideoX-I2V
& $1.60 \pm 0.39$
& $1.58 \pm 0.51$
& $1.47 \pm 0.36$
& $1.28 \pm 0.23$
& $1.36 \pm 0.25$
& 37.4
& 2.49 & 1.79 & 2.27 & 1.27 & 1.92 & 14.0 \\

HunyuanVideo-I2V
& $2.54 \pm 1.04$
& $2.18 \pm 1.05$
& $1.99 \pm 0.79$
& $1.73 \pm 0.53$
& $1.85 \pm 0.67$
& 16.2
& 2.88 & 2.00 & 2.71 & 1.57 & 2.16 & 10.0 \\

Wan-I2V
& $2.19 \pm 0.58$
& $1.96 \pm 0.63$
& $1.85 \pm 0.50$
& $1.70 \pm 0.47$
& $1.75 \pm 0.47$
& 30.4
& 3.33 & 2.69 & 3.43 & 2.11 & 2.90 & 0.0 \\

LTX-Video
& $2.52 \pm 0.84$
& $1.95 \pm 0.67$
& $2.11 \pm 0.74$
& $1.83 \pm 0.58$
& $1.95 \pm 0.59$
& 10.5
& 3.32 & 2.73 & 3.43 & 2.12 & 2.87 & 0.0 \\

\midrule


\multicolumn{13}{l}{
\textit{Physics-Grounded Methods}
} \\

PhysGen
& $3.55 \pm 0.58$
& $3.58 \pm 0.71$
& $2.51 \pm 0.57$
& $3.39 \pm 0.89$
& $3.06 \pm 0.67$
& 0.4
& 3.45 & 2.98 & 3.58 & 2.35 & 3.05 & 0.0 \\

MotionCraft
& $1.33 \pm 0.35$
& $1.26 \pm 0.20$
& $1.26 \pm 0.30$
& $1.11 \pm 0.10$
& $1.26 \pm 0.32$
& 55.5
& 1.96 & 1.48 & 1.37 & 1.07 & 1.54 & 14.0 \\

\midrule


\multicolumn{13}{l}{
\textit{Motion-Controllable Methods}
} \\

MotionPro
& $1.67 \pm 0.40$
& $1.72 \pm 0.66$
& $1.38 \pm 0.38$
& $1.28 \pm 0.24$
& $1.34 \pm 0.29$
& 35.2
& 2.65 & 1.85 & 2.25 & 1.35 & 1.75 & 16.0 \\

WanMove
& $3.38 \pm 0.86$
& $3.07 \pm 1.01$
& $2.88 \pm 0.88$
& $2.73 \pm 0.80$
& $2.81 \pm 0.83$
& 6.2
& 3.10 & 2.42 & 2.95 & 1.87 & 2.45 & 3.0 \\

\midrule


\textbf{IMPACT (Ours)}
& $\mathbf{3.87 \pm 0.61}$
& $\mathbf{3.69 \pm 0.34}$
& $\mathbf{3.72 \pm 0.55}$
& $\mathbf{3.47 \pm 0.64}$
& $\mathbf{3.61 \pm 0.49}$
& \textbf{0.0}
& \textbf{4.25}
& \textbf{3.69}
& \textbf{4.16}
& \textbf{3.40}
& \textbf{3.58}
& 0.0 \\

\bottomrule
\end{tabular}%
}
\end{table*}

%% file: Sections/4-experiments.tex
\section{Experiments}
\label{sec:experiment}

\subsection{Experimental Setup}
\label{sec:experimental_setup}

\noindent\textbf{Benchmarks.}
Two datasets are used to evaluate video generation and physical-state prediction.
Specifically, we construct a rigid-body collision benchmark
containing 100 initial images and corresponding text prompts.
The benchmark covers six sphere--ground material pairs
(rubber--wood, rubber--grass, stone--grass, stone--cement,
steel--cement, and steel--grass) across diverse scene
backgrounds. For each material--scene setting, we
systematically vary the initial release height, camera-relative
depth, and sphere size, together with object appearance and
illumination conditions. This design provides diverse physical
initial conditions and visual configurations, enabling a
comprehensive evaluation of physical motion generation
across varying geometric and environmental settings.
For physical-state evaluation, we construct a Kubric-simulated dataset containing 18,900 rigid-body collision sequences with reference trajectories and collision events. The dataset comprises 17,010 training sequences and 1,890 validation sequences, with the latter used for physical-state assessment.

\noindent\textbf{Implementation Details.}
All experiments are conducted on a computing server
equipped with four NVIDIA A100 GPUs.
Our method generates 81-frame videos at 25~fps
with a resolution of $1024\times512$ under a
fixed-camera setting.
The pretrained LTX-Video backbone and trajectory
conditioning module remain frozen, with optimization
restricted to the lightweight physical parameter
calibration components of PSG.

%

%
\noindent\textbf{Comparison Methods.}
We consider three categories of competing methods:
general-purpose image-to-video generators
(CogVideoX-I2V~\cite{cogvideox},
HunyuanVideo-I2V~\cite{hunyuanvideo},
Wan-I2V~\cite{wan}, and
LTX-Video~\cite{hacohen2025ltxvideo}),
physics-grounded methods (PhysGen~\cite{physgen} and MotionCraft~\cite{motioncraft}),
and motion-controllable methods (MotionPro~\cite{motionpro} and WanMove~\cite{wanmove}).

\cref{tab:inference_config} summarizes the
inference configurations and additional input
requirements of these methods.
All methods are evaluated using identical initial
images and text prompts, with consistent generation
settings where supported; otherwise, we follow
the officially recommended configurations.
Unlike methods requiring externally specified
physical parameters or motion controls,
our framework automatically derives physical
priors from the initial image and text prompt
without additional user-provided inputs.
Each method generates one video per input
without best-of-$N$ selection.

\noindent\textbf{Evaluation Metrics.}
We evaluate IMPACT from three complementary perspectives: human evaluation, vision-language-model (VLM) evaluation, and objective physical evaluation. Human and VLM assessments measure the perceived physical plausibility of generated videos, while objective physical metrics quantify the accuracy of predicted state evolution against simulation references.

\textbf{Human and VLM Evaluation.}
Both evaluations adopt five criteria: Physical Event Completeness (PEC), Free-Fall Dynamics Realism (FDR), Contact and Collision Realism (CCR), Rebound and Energy Dissipation Consistency (REC), and Overall Physical Plausibility (OPP). Each criterion is rated on a five-point scale, with higher scores indicating greater physical plausibility. Ten graduate students independently assess anonymized videos, and the reported scores are averaged across videos and raters. For VLM evaluation, 16 uniformly sampled frames from each video are assessed alongside the corresponding text and material conditions using the same criteria. We additionally report the Complete Failure Rate (CFR), defined as the proportion of evaluation units receiving the minimum score across all five criteria. Human CFR is computed over video--rater pairs, whereas VLM CFR is computed over videos.

\textbf{Objective Physical Evaluation.}
To complement perceptual assessments, we employ five objective metrics to quantify deviations from simulation references in continuous state evolution and discrete collision dynamics.

\textit{Continuous-State Accuracy.}
Position Error (PE) and Velocity Error (VE) measure the average Euclidean distances between predicted and reference positions and velocities, respectively:
\begin{equation}
\begin{aligned}
\mathrm{PE}
&=\frac{1}{N}\sum_{i=1}^{N}\frac{1}{T_i}
\sum_{t=1}^{T_i}
\left\|\hat{\mathbf p}_{i,t}
-\mathbf p_{i,t}^{*}\right\|_2,\\
\mathrm{VE}
&=\frac{1}{N}\sum_{i=1}^{N}\frac{1}{T_i}
\sum_{t=1}^{T_i}
\left\|\hat{\mathbf v}_{i,t}
-\mathbf v_{i,t}^{*}\right\|_2,
\end{aligned}
\label{eq:state_errors}
\end{equation}
where $N$ denotes the number of evaluation sequences and $T_i$ is the number of evaluated time steps in sequence $i$. The superscripts $\hat{\cdot}$ and $*$ indicate predicted and reference quantities, respectively.

\textit{Collision-Event Accuracy.}
To evaluate discrete collision dynamics, predicted impacts are matched one-to-one with reference impacts within a predefined temporal tolerance. Impact-detection F1 measures collision-event detection accuracy:
\begin{equation}
\mathrm{F1}
=\frac{2\mathrm{TP}}
{2\mathrm{TP}+\mathrm{FP}+\mathrm{FN}},
\label{eq:impact_f1}
\end{equation}
where TP, FP, and FN denote correctly matched impacts, unmatched predicted impacts, and missed reference impacts, respectively.

For matched events, Impact-Time Error (TE) and Rebound-Apex Error (AE) measure the mean absolute deviations in collision timing and maximum rebound height, respectively:
\begin{equation}
\begin{aligned}
\mathrm{TE}
&=\frac{1}{|\mathcal M|}
\sum_{(i,j,k)\in\mathcal M}
\left|\hat{\tau}_{i,j}-\tau_{i,k}^{*}\right|,\\
\mathrm{AE}
&=\frac{1}{|\mathcal M_{\mathrm a}|}
\sum_{(i,j,k)\in\mathcal M_{\mathrm a}}
\left|\hat{h}_{i,j}-h_{i,k}^{*}\right|,
\end{aligned}
\label{eq:impact_errors}
\end{equation}
where $\mathcal M$ denotes the set of matched impact pairs and $\mathcal M_{\mathrm a}\subseteq\mathcal M$ contains those with observable rebound apexes. Here, $\tau$ denotes the impact timestamp, and $h$ represents the maximum post-impact height of the sphere's lowest point above the ground plane. Since TE and AE are computed only for matched events, they are interpreted jointly with F1 to account for missed and spurious collisions.

\subsection{Performance Comparison}
\label{sec:performance_comparison}

\subsubsection{Overall Quantitative Comparison}
\label{sec:overall_quantitative_comparison}

We evaluate our method against three categories of
baselines: general-purpose video generators
(CogVideoX-I2V, HunyuanVideo-I2V, Wan-I2V, and
LTX-Video), physics-grounded methods (PhysGen and
MotionCraft), and motion-controllable methods
(MotionPro and WanMove).

As shown in \cref{tab:overall_comparison},
our method consistently outperforms all competing
approaches across the evaluation criteria while
achieving the lowest complete-failure rate.
Notably, the discrepancy between failure rates
and individual criterion scores indicates that
avoiding complete generation failures does not
necessarily ensure physical consistency.
These results highlight the limitations of
existing approaches in maintaining coherent
physical dynamics throughout video generation.

To further examine human-perceived quality,
we conduct pairwise comparisons based on
Overall Preference (OPP) ratings.
The winning fractions in
Fig.~\ref{fig:opp_winrate}
show that our method is consistently preferred
over competing approaches.
This preference-level evidence complements
the aggregate evaluation scores and further
demonstrates the perceptual advantages of
physics-guided generation.

\subsubsection{Comparison with General-Purpose Video Generators}
\label{sec:general_purpose_comparison}

General-purpose video generators rely primarily
on learned visual and temporal correlations,
without explicitly enforcing physical state
transitions.
Fig.~\ref{fig:qualitative_comparison} presents
representative results for outdoor sphere--grass
and indoor sphere--floor interactions.

CogVideoX-I2V exhibits visual artifacts and
object disappearance, disrupting appearance
consistency and motion continuity.
Wan-I2V shows implausible floating and fails
to settle, while HunyuanVideo-I2V produces
persistent oscillations with insufficient
rebound attenuation.
LTX-Video exhibits motion stagnation,
interrupting the progression of the
bouncing process.

In contrast, our method preserves a coherent
sequence of ground impact, progressively
attenuated rebounds, and eventual rest,
while maintaining stable object appearance.
These observations demonstrate the importance
of explicitly modeling physical events
rather than relying solely on learned
motion patterns.

\subsubsection{Comparison with Physics-Grounded Methods}
\label{sec:physics_grounded_comparison}

We compare our method with PhysGen and MotionCraft
to investigate whether physics-derived motion
can be faithfully translated into generated
video content.

PhysGen employs vision-language-based physical
property estimation and user-specified motion
conditions to construct image-space simulation
and motion references.
However, in our evaluated examples, it exhibits
inaccurate contact timing and misaligned
contact locations.
Its generated motion often resembles the
translation of a rigid two-dimensional cutout,
with limited natural appearance variation
and imperfect scene integration.

MotionCraft incorporates simulation-derived
optical flow through latent noise warping.
Nevertheless, it frequently fails to preserve
the complete progression from ground impact
to damped rebounds and eventual rest,
occasionally introducing object duplication
and visual artifacts.

\begin{figure*}[t]
\centering
\includegraphics[width=0.96\textwidth]
{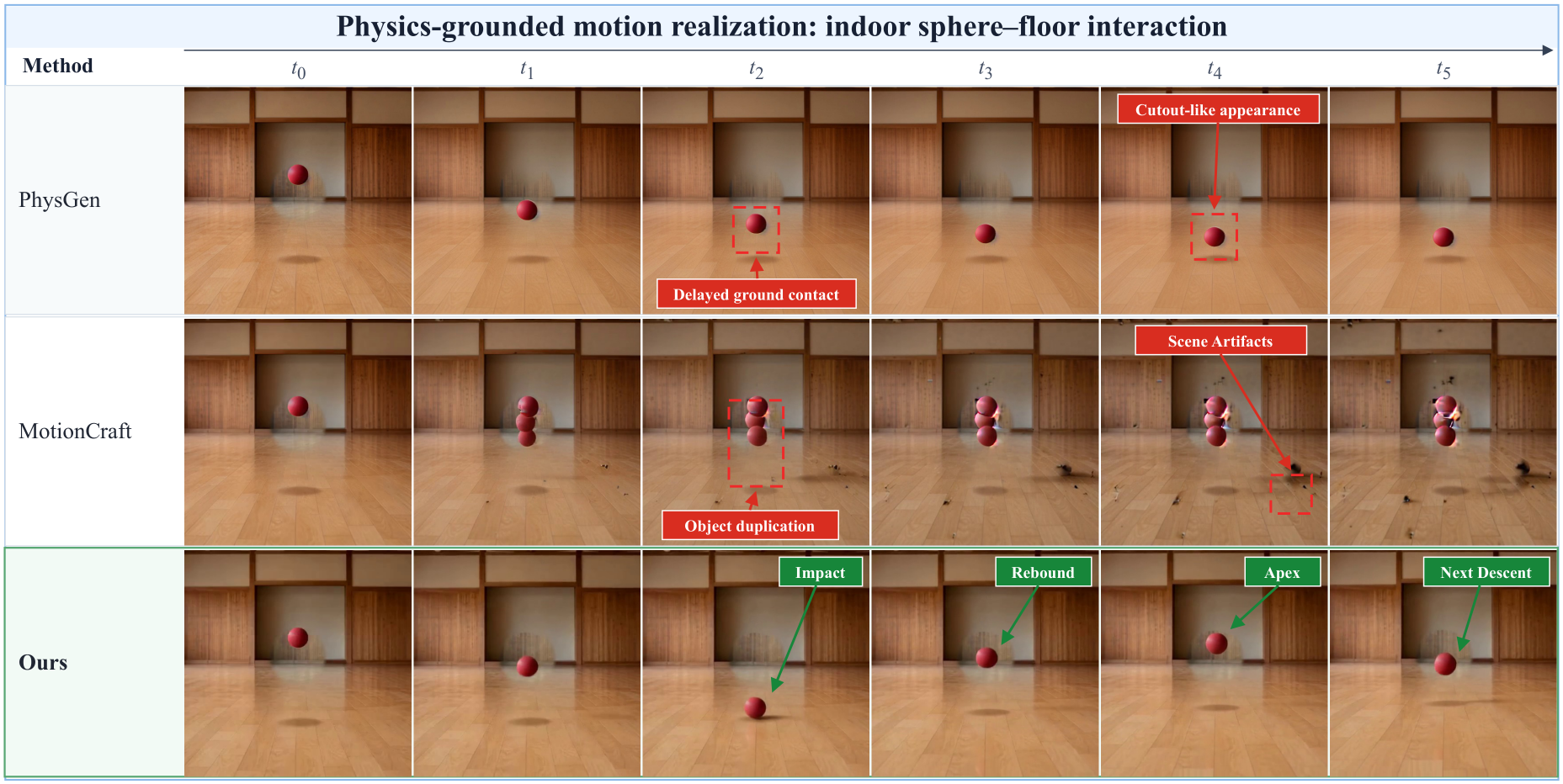}
\vspace{-0.2cm}
\caption{
Qualitative comparison with physics-grounded methods
at matched timestamps.
PhysGen exhibits delayed ground contact and
cutout-like motion, while MotionCraft suffers
from object duplication and scene artifacts.
In contrast, our method produces coherent
ground contact and rebound dynamics while
maintaining visual consistency.
}
\label{fig:physics_grounded_comparison}
\end{figure*}

As illustrated in
Fig.~\ref{fig:physics_grounded_comparison},
our method produces more coherent sphere--ground
interactions while preserving natural object
appearance and scene integration.
Unlike PhysGen, our framework automatically
constructs motion-relevant physical states
from a single image and text prompt without
requiring user-specified initial forces or
velocities for the considered free-fall task.
Moreover, our framework incorporates
physics-derived constraints directly into
video diffusion rather than relying on
pre-warped image sequences.
This comparison highlights the importance
of coupling explicit physical simulation
with generation-time guidance.

\subsubsection{Comparison with Motion-Controllable Methods}
\label{sec:motion_control_comparison}

We compare our method with MotionPro and WanMove
to examine whether physically consistent motion
inputs are sufficient to ensure coherent
sphere--ground interactions.
MotionPro employs region-wise trajectories
and motion masks, while WanMove introduces
latent trajectory guidance.
For a controlled comparison, we convert
the motion priors generated by our PSG into
the respective control representations
required by both methods.
Thus, all methods receive motion constraints
derived from the same physical state evolution.

Despite these physically grounded inputs,
MotionPro often produces limited object
displacement and fails to establish a complete
physical event sequence.
The sphere may disappear near or after
ground contact, disrupting appearance
continuity and rebound behavior.
WanMove produces more noticeable motion but
frequently exhibits nearly uniform falling
speed, unphysical floating, and persistent
movement without eventual settling.

\begin{figure*}[t]
\centering
\includegraphics[width=0.96\textwidth]
{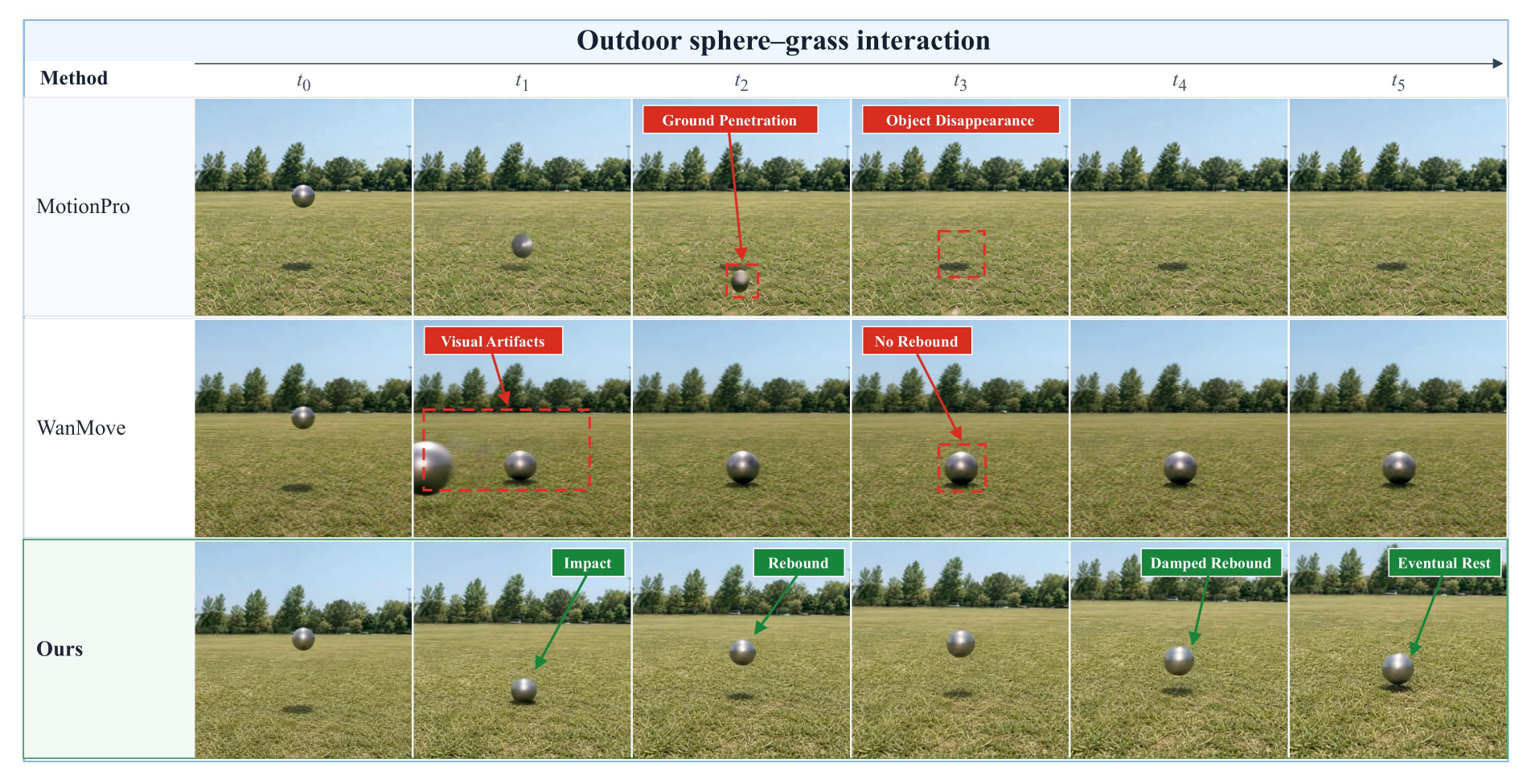}
\vspace{-0.4cm}
\caption{
Qualitative comparison with motion-controllable
methods at matched timestamps.
MotionPro exhibits ground penetration and
object disappearance, while WanMove suffers
from visual artifacts and missing rebound
behavior.
In contrast, our method produces coherent
ground contact, damped rebounds, and
eventual rest.
}
\label{fig:motion_controllable_comparison}
\end{figure*}

As illustrated in
Fig.~\ref{fig:motion_controllable_comparison},
our method achieves more complete physical
event progression while maintaining stable
object appearance.
These results suggest that physically
consistent trajectory inputs alone cannot
guarantee faithful physical motion execution.
By combining sparse trajectory conditioning
with dense physics guidance, our framework
more effectively translates the PSG-derived
physical states into temporally coherent
video dynamics.

%
\begin{table}[t]
\centering
\caption{
Physical-state prediction accuracy on the Kubric
validation set under identical I2P initialization.
}
\label{tab:psg_accuracy}
\footnotesize
\setlength{\tabcolsep}{3pt}
\renewcommand{\arraystretch}{1.15}
\begin{tabular*}{\columnwidth}
{@{\extracolsep{\fill}}lccccc@{}}
\toprule
\textbf{Method}
& \textbf{PE}$\downarrow$
& \textbf{VE}$\downarrow$
& \textbf{F1}$\uparrow$
& \textbf{TE}$\downarrow$
& \textbf{AE}$\downarrow$ \\
\midrule

Analytic Physics
& 0.5249 & 1.4353 & 0.372 & 0.0333 & 0.7746 \\

\textbf{PSG (Full)}
& \textbf{0.2658}
& \textbf{1.1685}
& \textbf{0.530}
& \textbf{0.0261}
& \textbf{0.1751} \\

\bottomrule
\end{tabular*}
\end{table}
%

\subsection{Physical State Prediction Accuracy}
\label{sec:state_accuracy}

\noindent\textbf{Evaluation Protocol.}
To assess the physical modeling capability of PSG
independently of visual estimation and video synthesis,
we conduct a controlled evaluation on a Kubric-based
validation set with simulation-derived ground truth.
For each sequence $i$, the simulator provides reference
sphere-center positions and velocities
$\{(\mathbf p_{i,t}^{*},\mathbf v_{i,t}^{*})\}_{t=1}^{T_i}$.
Given the simulation-defined initial conditions,
PSG predicts the corresponding state trajectory
$\{(\hat{\mathbf p}_{i,t},\hat{\mathbf v}_{i,t})\}_{t=1}^{T_i}$.
All trajectories are evaluated in a common physical
coordinate system at identical time steps, enabling
direct assessment of physical-state evolution.

To examine the effectiveness of PSG's dynamics modeling,
we compare it with an analytical physics baseline
that describes free fall under constant gravitational
acceleration and collisions using fixed
material-dependent restitution coefficients.
In contrast, PSG incorporates learned gravity
calibration and impact-velocity-dependent restitution
to model physical state transitions.
Both methods share identical simulation-defined
initial states, material configurations, ground geometry,
and prediction horizons, with reference future states
reserved exclusively for evaluation.
By eliminating variations in initialization and scene
configuration, this protocol isolates the effects
of dynamics modeling and enables a controlled
comparison of their ability to reproduce
free-fall, impact, and rebound dynamics.

\noindent\textbf{Quantitative Results.}
As shown in \cref{tab:psg_accuracy}, PSG outperforms
the analytic baseline across all five metrics,
demonstrating closer agreement with the Kubric
simulation references.
For continuous state prediction, PSG reduces position
and velocity errors by 49.4\% and 18.6\%,
respectively, indicating more accurate motion evolution.
The improvements extend to discrete collision dynamics:
impact-detection F1 increases from 0.372 to 0.530,
while impact-time error decreases by 21.6\%.
Notably, PSG reduces rebound-apex error by 77.4\%,
demonstrating substantially improved prediction of
post-impact rebound amplitudes.
These results indicate that PSG more faithfully
reproduces both continuous motion and collision-induced
state transitions than the fixed-restitution analytic
model, providing more accurate physical state sequences
for subsequent video generation.

%
\begin{table}[t]
\centering
\caption{
Impact-supervision ablation and initial-state comparison
on the Kubric validation set.
}
\label{tab:psg_ablation}
\footnotesize
\setlength{\tabcolsep}{3pt}
\renewcommand{\arraystretch}{1.15}
\begin{tabular*}{\columnwidth}
{@{\extracolsep{\fill}}lccccc@{}}
\toprule
\textbf{Variant}
& \textbf{PE}$\downarrow$
& \textbf{VE}$\downarrow$
& \textbf{F1}$\uparrow$
& \textbf{TE}$\downarrow$
& \textbf{AE}$\downarrow$ \\
\midrule

w/o Impact Loss
& 0.2677 & 1.6037 & 0.477 & 0.0626 & 0.1769 \\

Full (Oracle)
& 0.2422 & 1.2131 & 0.537 & 0.0529 & 0.1573 \\

Full (I2P)
& 0.2658 & 1.1685 & 0.530 & 0.0261 & 0.1751 \\

\bottomrule
\end{tabular*}
\end{table}

\begin{table*}[t]
\centering
\caption{
Ablation of motion-prior construction and
sparse--dense physics guidance under human
and VLM evaluations.
Human evaluation scores are reported as the mean
$\pm$ standard deviation of evaluator-level mean
ratings on a 1-5 scale.
}
\label{tab:motion_prior_ablation}

\footnotesize
\setlength{\tabcolsep}{4pt}
\renewcommand{\arraystretch}{1.12}

\resizebox{\textwidth}{!}{%
\begin{tabular}{l*{12}{c}}
\toprule

\multirow{2}{*}{\textbf{Method}}
& \multicolumn{6}{c}{\textbf{Human Evaluation}}
& \multicolumn{6}{c}{\textbf{VLM Evaluation}} \\

\cmidrule(lr){2-7}
\cmidrule(lr){8-13}

& PEC$\uparrow$
& FDR$\uparrow$
& CCR$\uparrow$
& REC$\uparrow$
& OPP$\uparrow$
& \mbox{CFR (\%)$\downarrow$}
& PEC$\uparrow$
& FDR$\uparrow$
& CCR$\uparrow$
& REC$\uparrow$
& OPP$\uparrow$
& \mbox{CFR (\%)$\downarrow$} \\

\midrule


\multicolumn{13}{l}{
\textit{Trajectory Variants}
} \\

PSG-KF Linear
& $3.46 \pm 0.61$
& $3.05 \pm 0.40$
& $3.22 \pm 0.40$
& $2.79 \pm 0.19$
& $2.95 \pm 0.23$
& 0.3
& 4.18 & 3.29 & 4.02 & 3.12 & 3.25 & 0.0 \\

PSG-KF Smoothstep
& $3.40 \pm 0.58$
& $3.06 \pm 0.41$
& $3.26 \pm 0.41$
& $2.74 \pm 0.44$
& $2.94 \pm 0.35$
& 0.3
& 4.24 & 3.62 & 4.13 & 3.34 & 3.51 & 0.0 \\

Naive Physics
& $3.33 \pm 0.76$
& $2.91 \pm 0.53$
& $3.28 \pm 0.53$
& $2.62 \pm 0.49$
& $2.86 \pm 0.38$
& 0.3
& 4.22 & 3.47 & 4.10 & 3.27 & 3.42 & 0.0 \\

\midrule


\multicolumn{13}{l}{
\textit{Sparse--Dense Guidance}
} \\

LTX-Video
& $2.52 \pm 0.84$
& $1.95 \pm 0.67$
& $2.11 \pm 0.74$
& $1.83 \pm 0.58$
& $1.95 \pm 0.59$
& 10.5
& 3.32 & 2.73 & 3.43 & 2.12 & 2.87 & 0.0 \\

PSG + Sparse
& $2.73 \pm 0.75$
& $2.34 \pm 0.53$
& $2.55 \pm 0.59$
& $2.22 \pm 0.42$
& $2.38 \pm 0.44$
& 4.8
& 3.25 & 2.54 & 3.17 & 2.34 & 2.51 & 0.0 \\

\midrule


\textbf{IMPACT (Ours)}
& $\mathbf{3.87 \pm 0.61}$
& $\mathbf{3.69 \pm 0.34}$
& $\mathbf{3.72 \pm 0.55}$
& $\mathbf{3.47 \pm 0.64}$
& $\mathbf{3.61 \pm 0.49}$
& \textbf{0.0}
& \textbf{4.25}
& \textbf{3.69}
& \textbf{4.16}
& \textbf{3.40}
& \textbf{3.58}
& 0.0 \\

\bottomrule
\end{tabular}%
}
\end{table*}
\begin{figure*}[t]
\centering
\IfFileExists{Figures/ablation_comparison.pdf}{%
  \includegraphics[width=\textwidth]
  {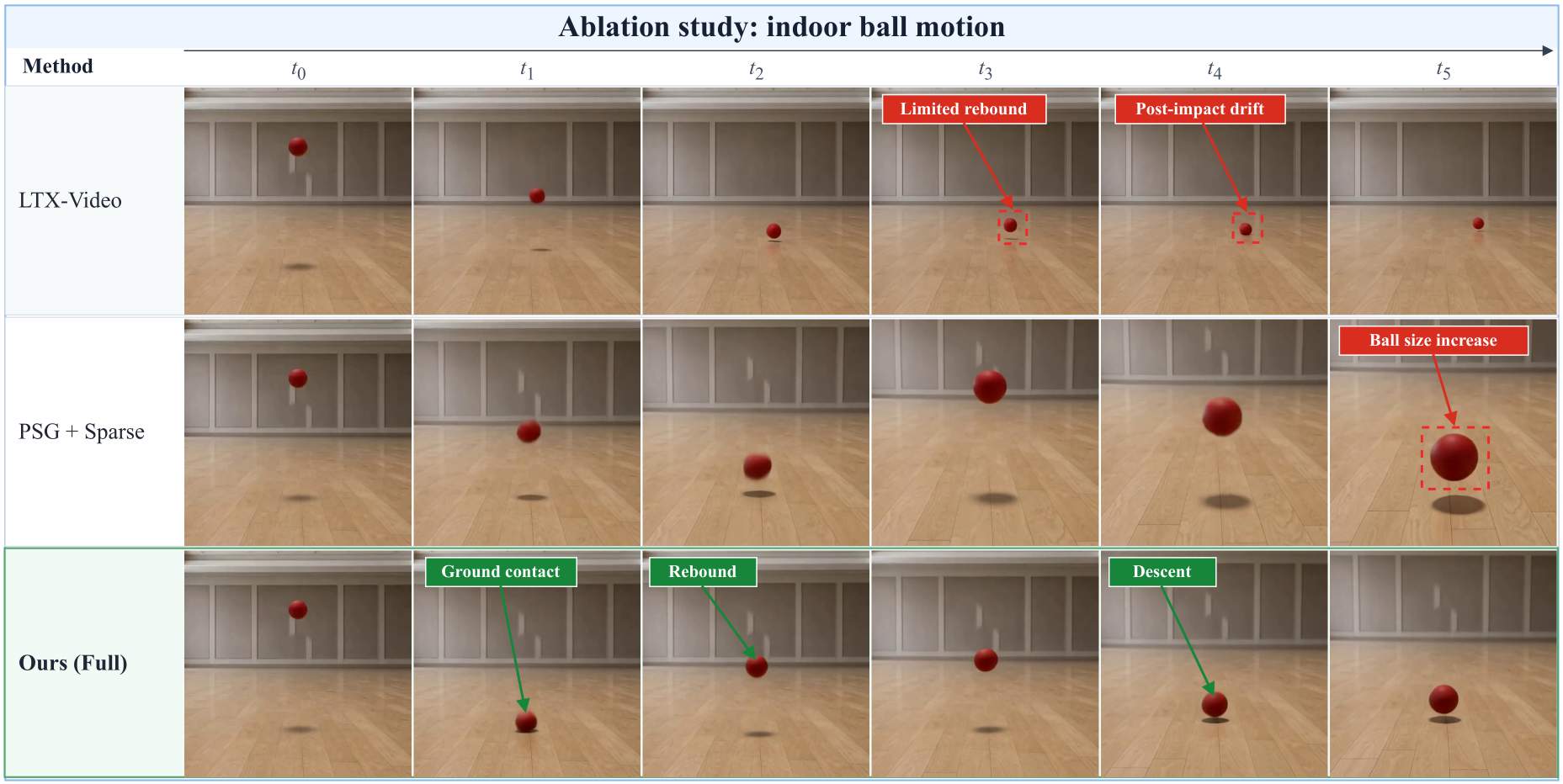}%
}{%
  \fbox{\parbox[c][1.1in][c]{0.91\textwidth}{%
  \centering Qualitative guidance ablation:
  Figures/ablation_comparison.pdf}}%
}
\vspace{-0.5cm}
\caption{
Qualitative comparison of LTX-Video, PSG + Sparse,
and Ours (Full) at matched timestamps.
LTX-Video exhibits spatial deviations and
physically inconsistent motion despite producing
ground collisions.
PSG + Sparse follows the predicted trajectory
but suffers from local object enlargement,
revealing a prediction--generation mismatch.
In contrast, our full method preserves more
consistent object scale and coherent collision dynamics.
}
\label{fig:guidance_ablation}
\end{figure*}

\subsection{Ablation Studies}
\label{sec:ablation_studies}
\noindent\textbf{Impact Supervision and State Initialization.}
We further investigate the contribution of impact supervision
and the influence of initial-state estimation on PSG.
For the supervision ablation, the full model and the variant
without $\mathcal L_{\mathrm{impact}}$ share identical oracle
initial states, training data, network architectures, and
optimization settings, differing only in the training
objective. We additionally evaluate the same full PSG
using I2P-estimated initial states to examine its
sensitivity to initialization.

As shown in \cref{tab:psg_ablation}, removing impact
supervision degrades performance across all evaluated metrics
under otherwise identical conditions. This demonstrates that
impact supervision complements trajectory-level position
constraints by improving the prediction of collision-induced
state transitions and subsequent rebound dynamics.

When oracle initialization is replaced with I2P estimates,
the full PSG maintains broadly comparable state-prediction
and impact-detection performance, although individual metrics
respond differently. This indicates that PSG can operate
with image-estimated physical states without requiring
simulator-provided initialization. Since impact-time errors
are evaluated only for matched events, their interpretation
should remain coupled with impact-detection F1.

\noindent\textbf{Motion-Prior Construction and Sparse--Dense Guidance.}
To investigate the influence of motion-prior construction,
we compare several trajectory variants and guidance settings,
as summarized in \cref{tab:motion_prior_ablation}.
\emph{PSG-KF Linear} and \emph{PSG-KF Smoothstep}
share the same PSG-derived event keyframes but employ
different inter-event interpolation strategies.
\emph{Naive Physics} simulates free fall and repeated
collisions using constant gravity and fixed
material-dependent restitution coefficients.
We further compare \emph{PSG + Sparse}, which uses
the complete PSG trajectory without dense guidance,
with \emph{Ours (Full)}, which additionally incorporates
dense physics guidance during diffusion sampling.

The results reveal that perceptually smooth motion
does not necessarily imply physically accurate state
transitions, highlighting the importance of
event-driven physical modeling.
Moreover, sparse trajectory conditioning alone
cannot fully enforce physical consistency during
video synthesis, motivating the integration of
dense generation-time guidance.

As illustrated in Fig.~\ref{fig:guidance_ablation},
LTX-Video produces ground collisions but exhibits
spatial deviations and physically implausible motion.
Although PSG + Sparse largely follows the
predicted trajectory, the generated sphere
undergoes local enlargement, revealing a
prediction--generation mismatch between
physically valid motion predictions and
their visual realization.
In contrast, Ours (Full) better maintains
object scale and coherent contact behavior
throughout the motion sequence.
These observations demonstrate that sparse
trajectory control primarily constrains global
displacement, whereas dense physics guidance
is essential for improving local physical
consistency during video generation.

\subsection{Limitation and Discussion}

The current framework is developed and evaluated under a controlled sphere-plane interaction setting, where a single rigid sphere is released from rest and undergoes successive collisions with a stationary plane under a fixed camera. More complex interactions, such as collisions between multiple objects, non-spherical geometries, rotational motion, and deformable-body dynamics, involve additional physical states and contact constraints that are not captured by the present formulation. Moreover, the initial physical configuration is estimated from a single image, making the predicted motion sensitive to uncertainties in depth, object geometry, and camera calibration. Extending PSG to more general physical interactions therefore requires richer state representations and more robust physical parameter estimation.

A further limitation concerns the translation of predicted physical states into generated video content. Although sparse-dense guidance improves motion realization, it does not guarantee exact agreement between the generated video and the underlying physical trajectory. The current dense priors primarily describe translational object motion and background stability, leaving fine-grained contact deformation, rotation, and complex appearance changes insufficiently constrained. In addition, sampling-time latent guidance introduces additional computational overhead and remains dependent on the representational capabilities of the pretrained video generator. These limitations suggest that physically consistent video generation requires not only accurate state prediction but also more expressive and efficient mechanisms for enforcing physical constraints during synthesis. Future work will explore richer physical representations and adaptive generation-time guidance to improve the fidelity and generality of physical motion realization.

%% file: Sections/5-conclusion.tex
\section{Conclusion}
\label{sec:conclusion}
In this paper, we presented IMPACT, a physics-state-driven framework for physically consistent image-to-video generation. Our findings highlight that physical consistency depends on both the coherent evolution of physical states and their faithful realization during video synthesis. Technically, IMPACT employs an event-driven Physics State Generator to connect continuous free-flight dynamics with calibrated collision responses, ensuring coherent transitions across successive physical events. It further translates the predicted physical evolution into complementary sparse trajectory and dense region-level constraints to guide motion realization during diffusion sampling. Experimental results demonstrate improved physical plausibility across different material combinations and visual environments. Notably, the ablation results reveal that trajectory guidance alone is insufficient to ensure physically consistent visual motion, highlighting the importance of region-level constraints during synthesis. These findings suggest that physically grounded video generation should move beyond plausible motion prediction toward a tighter integration of physical dynamics and generative control. Extending this principle to more complex physical interactions remains an important direction for future research.